\documentclass[10pt,twocolumn,letterpaper]{article}

\usepackage[pagenumbers]{cvpr} % To force page numbers, e.g. for an arXiv version

\usepackage{graphicx}
\usepackage{amsmath}
\usepackage{amssymb}
\usepackage{booktabs}
\usepackage{comment}
\usepackage{color}
\usepackage{multirow}
\usepackage{gensymb}
\usepackage{pifont}
\usepackage{enumerate}

\usepackage[pagebackref=true,breaklinks=true,letterpaper=true,colorlinks,bookmarks=false]{hyperref}
\usepackage[capitalize]{cleveref}
\crefname{section}{Sec.}{Secs.}
\Crefname{section}{Section}{Sections}
\Crefname{table}{Table}{Tables}
\crefname{table}{Tab.}{Tabs.}

\DeclareMathOperator*{\argmax}{arg\,max}
\definecolor{hlrowcolor}{rgb}{1.0,0.9,0.8}

\newcommand\blfootnote[1]{%
  \begingroup
  \renewcommand\thefootnote{}\footnote{#1}%
  \addtocounter{footnote}{-1}%
  \endgroup
}

\def\cvprPaperID{1956} % *** Enter the CVPR Paper ID here
\def\confName{CVPR}
\def\confYear{2023}

\begin{document}

\title{Accidental Turntables: Learning 3D Pose by Watching Objects Turn}

\author{Zezhou Cheng$^1$ \quad Matheus Gadelha$^2$ \quad Subhransu Maji$^1$\\
$^1$University of Massachusetts, Amherst \quad $^2$Adobe Research\\
{\tt\small \{zezhoucheng, smaji\}@cs.umass.edu \quad gadelha@adobe.com}
}
\maketitle

%%%%%%%%% ABSTRACT
\begin{abstract}
We propose a technique for learning single-view 3D object pose estimation models by utilizing a new source of data --- in-the-wild videos where objects turn.  
Such videos are prevalent in practice (\eg cars in roundabouts, airplanes near runways) and easy to collect.
We show that classical structure-from-motion algorithms, coupled with the recent advances in instance detection and feature matching, provides surprisingly accurate relative 3D pose estimation on such videos. 
We propose a multi-stage training scheme that first learns a canonical pose across a collection of videos and then supervises a model for single-view pose estimation.
The proposed technique achieves competitive performance with respect to existing state-of-the-art on standard benchmarks for 3D pose estimation, without requiring any pose labels during training. We also contribute an Accidental Turntables Dataset, containing a challenging set of 41,212 images of cars in cluttered backgrounds, motion blur and illumination changes that serves as a benchmark for 3D pose estimation.\blfootnote{Project website: \url{https://people.cs.umass.edu/~zezhoucheng/acci-turn/}}
\end{abstract}

\section{Introduction}
\begin{figure}
\centering
\includegraphics[width=\linewidth]{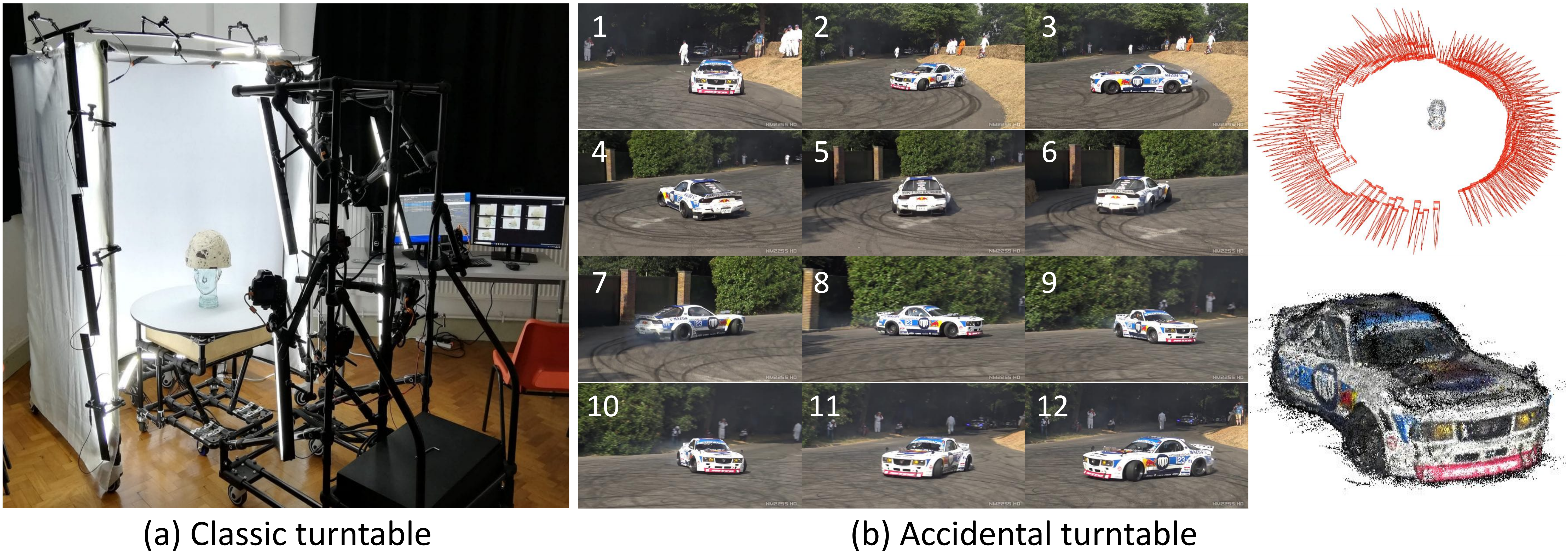}
% \vspace{-2mm}
\caption{\textbf{Classic turntable vs. accidental turntable.} \textbf{(a)} Classic turntables rotate and scan objects in a controlled environment for estimating their 3D pose and shape. \textbf{(b)} A turning object in a video leads to an accidental turntable. Structure-from-motion, coupled with object detection~\cite{he2017mask} and feature matching~\cite{sarlin2020superglue}, provides surprisingly accurate relative 3D pose estimation (top) and 3D reconstruction (bottom) --- the red pyramids indicate the estimated relative poses of video frames. We utilize a collection of such videos to train and evaluate models for single-frame 3D pose estimation in realistic settings. See more accidental turntables here: \url{https://www.youtube.com/watch?v=8rFNRri8-TI}}
\label{fig:turntable}
\end{figure}

Understanding object pose, and its structure is a central computer vision problem. 
Many images have been manually annotated with pose information in multiple datasets containing various types of objects.
Still, this manual annotation process is labor-intensive and prone to unavoidable human annotation errors.
On the other hand, mechanical devices that precisely change an object pose are widely utilized when performing high-precision 3D scanning.
They allow a particular object to have its pose modified in a controlled manner while capturing its appearance
through a variety of image sensors.
One of the simplest devices of this kind is a \emph{turntable} -- a rotating platform that slowly changes the
pose of an object through an electric motor (Fig.~\ref{fig:turntable}a). 
Unfortunately, despite its simplicity, \emph{turntables} are not very practical.
They need to be as large as the object at hand, \eg, setting up turntables for cars or airplanes would require a lot of work.

Fortunately, we don't need to place those objects in actual turntables.
Many are already performing similar motion on their own (Fig.~\ref{fig:turntable}b) --- cars moving along
roundabouts, airplanes landing and parking, ships maneuvering across canals, and so on.
In the real world, video recordings of objects performing these types of motions depict them in uncontrolled environments; \ie cluttered background, occluders, changes in illuminations, motion blur, unpredictable pose changes and many other nuisance factors.
Thanks to many recent advances in computer vision, we show that we are able to bypass many of those
nuisance factors and apply Structure from Motion (SfM) to reliably and precisely recover relative pose estimation from videos of real objects (Fig.~\ref{fig:turntable}b).
We call these types of videos \textbf{Accidental Turntables} -- objects presenting motion patterns that allow us to
observe them from (almost) all possible angles.
We demonstrate that these videos, after suitable automatic pre-processing (Sec.~\ref{sec:turntabledataset}), are an excellent source of supervision
for pose estimation models and, perhaps more importantly, can be mined from the internet, enabling the creation of bigger and more diverse datasets.

However, using the supervision from SfM does not allow us to directly perform pose estimation with respect to a canonical object frame.
To this end, we propose to learn a \emph{relative} pose estimation model and show that its training leads to the emergence of a canonical object pose (\S~\ref{sec:relative}).
In a second stage, we propose a calibration and training procedure (\S~\ref{sec:cano}) that allows pose estimation in a canonical frame (\S~\ref{sec:regression}).
We show that models trained in this fashion \emph{only} using our newly collected dataset from \emph{real videos} significantly outperforms other models trained on SfM and performs on par with existing unsupervised approaches on standard benchmarks, \eg, the Frieburg and ImageNet cars datasets. 

We summarize our contributions as follows. 
1) a procedure for automatically processing accidental turntable videos and annotating its frames with relative
pose transformations;
2) a multi-stage training scheme that allows training accurate pose estimation models with respect to
arbitrary canonical frames; and
3) a new dataset with 41,212 real images of cars from turntable videos with their corresponding pose annotation.

\section{Related Work}

\paragraph{\textbf{Datasets for pose estimation.}}   
A number of datasets provide 3D pose annotations for objects in the wild~\cite{ahmadyan2021objectron,song2019apollocar3d,sun2018pix3d,xiang2014beyond,xiang2016objectnet3d,geiger2012we} or in controlled environments~\cite{fang2020graspnet,wang2019normalized,wen2020se,xiang2017posecnn,hodan2017t}.
These datasets have been widely used for training supervised pose estimation models~\cite{tulsiani2015viewpoints,mahendran20173d,liao2019spherical,grabner20183d}.
However, manually annotating 3D pose is very tedious and thus not scalable. 
Unsupervised pose estimation models~\cite{sedaghat2015unsupervised,novotny2017learning,mustikovela2020self,mariotti2021viewnet} learn to predict 3D pose without any human annotations.
Videos~\cite{sedaghat2015unsupervised,ozuysal2009pose} that capture multiple views of objects have been the main source of training data in prior works~\cite{sedaghat2015unsupervised,novotny2017learning,mariotti2021viewnet}.
However, to acquire such videos, a person needs to hold a camera and slowly moves around a \emph{static} object. 
This is a time-consuming procedure, especially for large-size objects (\eg cars, airplanes), and has limits the size of existing video datasets.
For example, the Freiburg Cars dataset~\cite{sedaghat2015unsupervised} consists of 52 car videos and EPFL car dataset~\cite{ozuysal2009pose} only provides 20 cars.
Such limited data may further constrain the performance of prior methods.
In this work, we identify a new source of data for unsupervised pose estimation -- videos where objects turn. 
The turning of objects (\eg, vehicles) is such a natural phenomenon in daily life that these videos are quite easy to collect. 
We build a new dataset consisting of 313 car videos with a total of 141,784 frames.
Our 3D pose annotations are generated by SfM~\cite{schonberger2016structure,schonberger2016pixelwise} enhanced by recent progress in object detection~\cite{he2017mask} and feature matching~\cite{sarlin2020superglue}.

\paragraph{\textbf{Supervised pose estimation.}}  
With groundtruth 3D pose annotations, supervised pose estimation works have been focusing on developing novel representations of 3D pose~\cite{zhou2019continuity,liao2019spherical,murphy2021implicit}, learning objectives~\cite{tulsiani2015viewpoints,liao2019spherical,xiao2019pose,xiao2021posecontrast}, or network architectures~\cite{esteves2018learning,esteves2020spin}. 
The difficulty in annotating 3D pose results in the scarcity of pose annotations. 
This issue is partially relieved by augmenting the existing datasets with synthetic data~\cite{su2015render}. 
The integration of pose estimation and object detection has been explored in the task of 3D object detection~\cite{geiger2012we,divon2018viewpoint}.
Our models are built upon the prior supervised learning methods~\cite{zhou2019continuity,xiao2019pose,xiao2021posecontrast}, but we use pose annotations automatically generated by SfM, instead of human annotations.

\paragraph{\textbf{Unsupervised pose estimation.}}  
Unsupervised pose estimation models learn 3D object pose without any human annotations.
Prior works are either based on analysis-by-synthesis~\cite{mustikovela2020self,mariotti2021viewnet} or SfM~\cite{sedaghat2015unsupervised,novotny2017learning}.
The analysis-by-synthesis frameworks train a pose estimation model via reconstructing the input images in a pose-aware manner. 
The SfM-based methods start by estimating the pose labels with SfM on videos that capture $360^\circ$ views of static objects.
However, SfM only provides relative pose estimations among video frames. The absolute pose estimations from SfM are not consistent across videos (\ie objects in a same pose from two videos may have quite different absolute pose representations). 
To tackle this issue, Sedaghat~\etal~\cite{sedaghat2015unsupervised} calibrate the SfM pose estimations via aligning 3D reconstructions of objects;
Novotny~\etal~\cite{novotny2017learning} train a model to estimate the relative pose and observe that canonical poses emerge in the models trained in this manner.
Similar to Novotny~\etal~\cite{novotny2017learning}, we train a model to estimate the relative pose from SfM (Sec.~\ref{sec:relative}).
Differently, we find that such training strategy is not sufficient to learn a high-quality pose predictor. 
We instead use the model trained in this way as a tool to calibrate the SfM estimations across videos (Sec.~\ref{sec:cano}), followed by training a novel pose estimator on the calibrated pose annotations (Sec.~\ref{sec:regression}).

\paragraph{\textbf{Accidental data in computer vision.}} 
Researchers have discovered interesting phenomena that occur accidentally but turns out to be useful for computer vision tasks in the literature. 
Torralba~\etal~\cite{torralba2012accidental} demonstrate that outdoor scenes can be recovered from accidental pinhole images.
Li~\etal~\cite{li2019learning} train a depth estimator on a collection of Internet videos of people imitating mannequins, \ie freezing in diverse pose. 
The depth information are obtained from SfM and multi-view stereo algorithms.
Similar to Li~\etal~\cite{li2019learning}, we train a pose estimation model on a collection of Internet videos and the 3D pose annotations are automatically generated from SfM.
Unlike people imitating mannequins, we only require the object turns in the video, which is a quite natural behavior in practice (\eg, car moving along roundabouts). 

\section{Accidental Turntable Dataset}
\label{sec:turntabledataset}

 In this section, we provide the details of our data collection and the generation of 3D pose annotations with SfM algorithms on our dataset. 
 We name the collected video dataset as \textbf{Accidental Turntable Dataset}, highlighting it connections to classic turntables (Fig.~\ref{fig:turntable}).
 
\begin{figure}
\centering
\includegraphics[width=\linewidth]{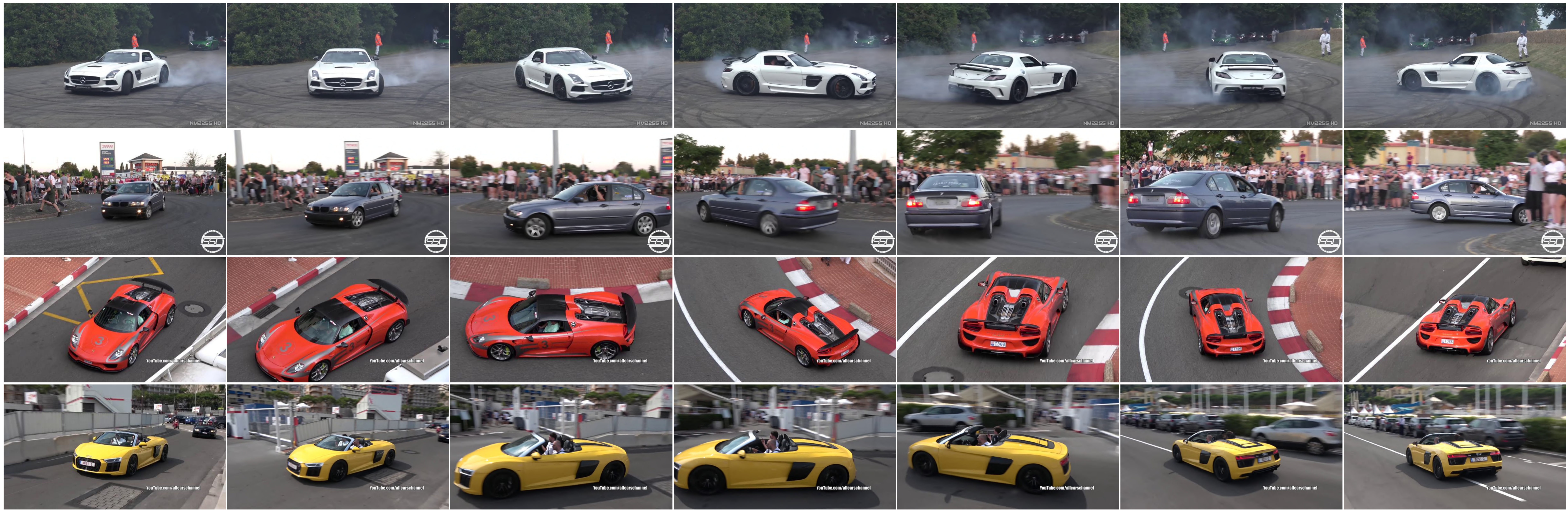}
\caption{\textbf{Samples from accidental turntable dataset.} The accidental turntables are prevalent in practice. For instance, a car donuts (1st row), a car moves along a roundabout (2nd and 3rd row), or a car does not turn but passes by a camera (4th row). All car instances exhibit at least 180$^\circ$ azimuth changes relative to the camera.}
\label{fig:video-example}
% \vspace{-5mm}
\end{figure}

\paragraph{\textbf{Data source.}} 
The main criterion of our data collection is that the object turns in the video.
Such videos are abundant on the Internet and quite easy to acquire. 
In this work, we focus on the car category which is one of the most common moving objects in the wild (at least in America). 
We leave the extension to other categories (\eg, airplanes and boats) in our future work, but include some examples of the reconstructions at the end of the paper.
We collect 313 car video clips from YouTube containing in total of 141,784 frames. 
Each video consists of a single moving car instance which exhibits multiple views in motion. 
Fig.~\ref{fig:video-example} provides video samples from our dataset. 
We provide a full list of YouTube links to the collected videos in Appendix~\ref{sec:more-dataset}. 

\paragraph{\textbf{Challenges.}}
Even though our dataset consists of large number of car videos serving as a new source training data for machine learning models, in-the-wild videos pose technical challenges for automatic extraction of 3D pose using SfM. 
For example, to exploit the classical SfM algorithms to estimate the object pose, object segmentation is required to remove the background; 
Motion blur and texture-free object surfaces necessitate robust interest points detection;
Discriminative feature description and robust feature matching are needed to avoid the ambiguity of pose estimation on symmetric objects (\eg, cars).

\paragraph{\textbf{Pose estimation with SfM.}} 
To tackle the above-mentioned challenges, we use the MaskRCNN~\cite{he2017mask} pretrained on MS-COCO dataset~\cite{lin2014microsoft} to remove the background clutter. 
We find that the MaskRCNN provides highly accurate object detection and segmentation on in-the-wild car videos.
We use SfM algorithms implemented by COLMAP~\cite{schonberger2016pixelwise,schonberger2016structure} with SuperPoint~\cite{detone2018superpoint} as the feature extractor and SuperGLUE~\cite{sarlin2020superglue} as the feature matcher to estimate the object pose on cropped object images. 
We sequentially match the next 10 frames per video frame, instead of exhaustive matching every pair of frames in a video.
The sequential matching reduces the ambiguity in matching repeated patterns (\eg left and right wheel of a car).
SfM, coupled with MaskRCNN, SuperPoint, and SuperGLUE, provides surprisingly accurate pose estimation, in comparison with classical SIFT~\cite{lowe2004distinctive} and nearest neighbor matching. 
We provide detailed study on the effect of feature extraction and matching on SfM in Sec.~\ref{sec:analysis}.

\paragraph{\textbf{Statistics.}} Our dataset consists of 313 car videos with 141,784 frames in total. 
SfM automatically samples frames with sufficient large relative pose change and reliable feature matching. 
Adjacent frames in a video usually have tiny difference in pose, thus most of frames are filtered out by SfM. 
We end up collecting 41,212 frames with SfM pose estimations. 
Our dataset covers cars with diverse shapes, colors, textures, and poses (see examples in Fig.~\ref{fig:video-example}). 

\section{Methodology}

\begin{figure}
\centering
\includegraphics[width=\linewidth]{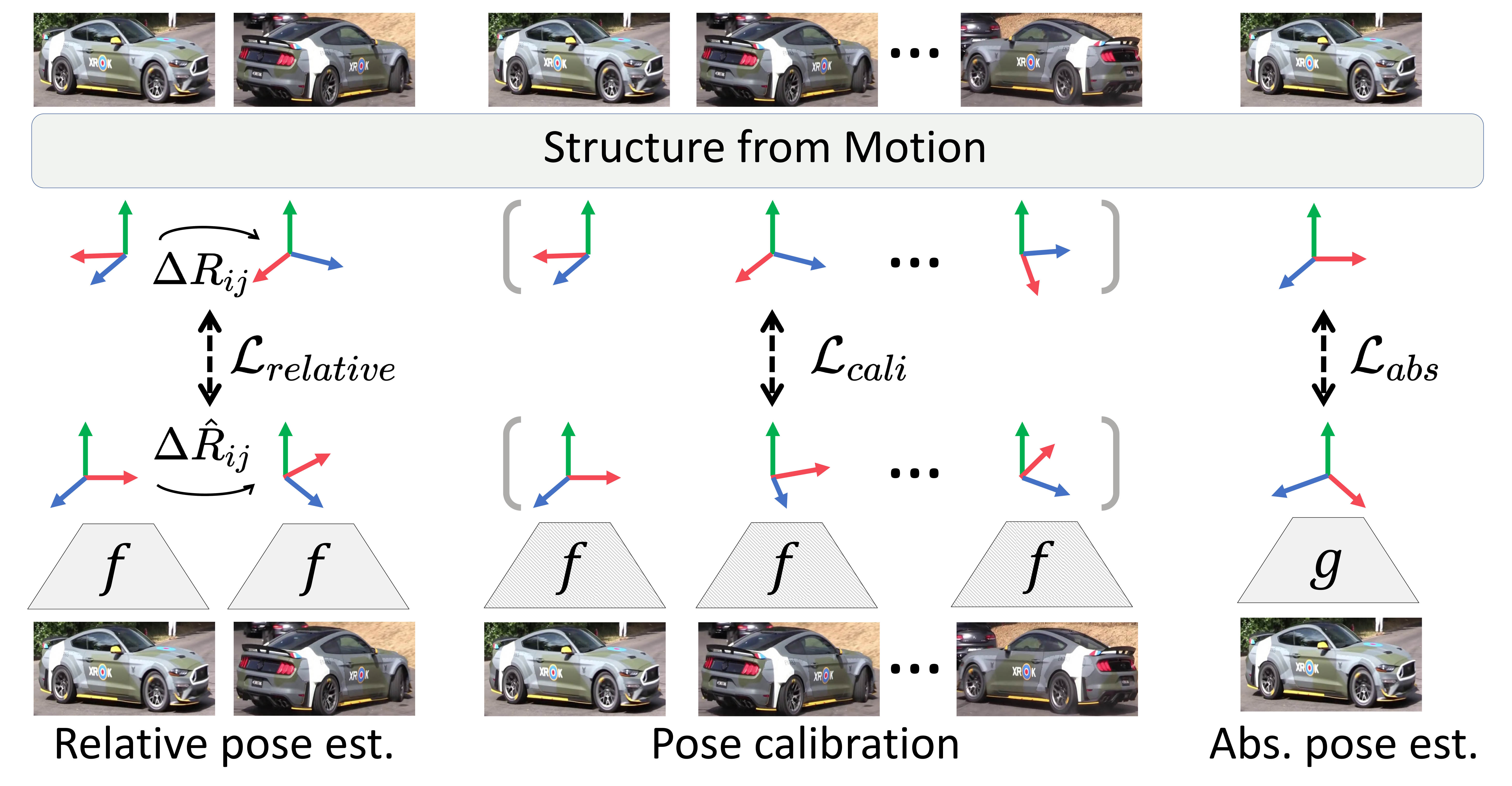}
\caption{\textbf{Approach overview.}
\textbf{Left:} a pose estimation model $f(x)$ is trained to predict the \emph{relative} pose of image pairs (denoted by $\Delta R_{ij}$).  
\textbf{Middle:} the emergence of canonical pose in $f(x)$ enables us to calibrate the pose estimations from SfM to a uniform frame. 
The model $f(x)$ is frozen in the pose calibration step.
\textbf{Right:} after the pose calibration, a pose estimation model $g(x)$ is trained on the \emph{absolute} pose annotations.}
\label{fig:overview}
% \vspace{-5mm}
\end{figure}

This section introduces our framework for learning 3D object pose from the proposed accidental turntable dataset.
Fig.~\ref{fig:overview} illustrates an overview of the proposed framework.
SfM estimates the relative pose of objects with respect to the object in the first frame per video, followed by optimizing the pose parameters with the bundle adjustment. 
However, the object pose in the first frame may vary dramatically across videos. 
It is thus meaningless to train a model directly on the absolute pose labels from SfM.
Instead, we start by training a model to estimate the \textit{relative} pose of frame pairs (Fig.~\ref{fig:overview} left). 
We observe that a canonical pose emerges in our pose estimation model train in this way (see Sec.~\ref{sec:analysis}). 
This provides us a tool to calibrate the pose estimation from SfM to a canonical frame  (Fig.~\ref{fig:overview} middle).
In the second stage, we train a pose estimation model directly on the calibrated \textit{absolute} pose annotations similar to standard supervised learning methods~\cite{tulsiani2015viewpoints,su2015render,xiao2021posecontrast}  (Fig.~\ref{fig:overview} right).
We denote our model trained in the first stage as $f(x)$ and the model in the second stage as $g(x)$, where $x$ is the input image.
Our accidental turntable dataset is denoted by $\{(x_i, R_i)\}$, where $R \in \text{SO(3)}$ is the SfM pose estimation.

\subsection{Relative pose estimation}\label{sec:relative}

In this stage, we train a single-view pose estimation network $f(x)$ to predict the relative pose between pairs of video frames. 
The loss function is defined as,  
\begin{align}
    \mathcal{L}_{\text{relative}} = \sum_{(i,j)}^N \text{dist}(R_iR_j^T, \hat{R}_i\hat{R}_j^T) \quad \text{with} \quad \hat{R}_i = f(x_i)
\end{align}
where $\text{dist}(\cdot, \cdot)$ is a distance function between two rotation matrices (\eg $L_2$ or geodesic distance). 
$\hat{R}_i$ is a $3\times 3$ rotation matrix predicted from the model $f(x_i)$ on the input $x_i$.
The frame pair $x_i$ and $x_j$ are sampled from a same video.
$N$ is the total number of frame pairs sampled from our video dataset. 
$\Delta R_{ij} = R_iR_j^T$ is the relative rotation matrix that transforms the pose of the frame $x_j$ to $x_i$. 
We use the 6D continuous rotation representation~\cite{zhou2019continuity} as intermediate output of our model $f(x)$, from which the $3 \times 3$ rotation matrices $\hat{R}$ are recovered by the Gram-Schmidt orthogonalization~\cite{zhou2019continuity}. 
Our first training stage is similar to the learning strategy proposed by Novotny~\etal~\cite{novotny2017learning}. 
Differently, we only use the model $f(x)$ trained in this stage as a tool to calibrate the SfM pose annotations (Sec.~\ref{sec:cano}).
Moreover, we demonstrate that the model $g(x)$ trained in our second stage significantly outperforms the stage-one model $f(x)$ as well as Novotny~\etal~\cite{novotny2017learning}. We provide detailed comparisons between $f(x)$ and $g(x)$ in Sec.~\ref{sec:analysis}.

\subsection{Pose calibration}\label{sec:cano}

The pose predictor $f(x)$ trained in the first stage provides us a tool to calibrate the pose annotations from SfM into a uniform pose frame, thanks to the emergence of canonical pose (see Sec.~\ref{sec:analysis} for more details).
If the pretrained $f(x)$ provides perfectly accurate pose estimation per input $x$, there exists a global rotation $\Delta R$ for each video that aligns our pose annotations $\{R_i\}$ to the pose predictions $\{\hat{R}_i\}$:
\begin{align}
 \hat{R}_i = \Delta RR_i \quad \forall i \in {1,\dots,K}
\end{align}
where $K$ is the number of frames in the target video. However, the pose predictions $\{\hat{R}_i\}$ are inaccurate in practice due to the limited performance of the pretrained pose predictor $f(x)$.
We thus target at a rotation matrix $\Delta R^*$ that aligns $\{R_i\}$ and $\{\hat{R}_i\}$ with minimal calibration error.
We define the calibration error as,
\begin{align}
 \mathcal{L}^*_{cali} = \frac{1}{K}\sum_i^{K} \text{dist}(\hat{R}_i, \Delta R^*R_i) \label{eq:alignerr} 
\end{align}
where $\text{dist}(\cdot, \cdot)$ is a distance function between two rotation matrices.
We adopt the geodesic distance $\|\log R^T\hat{R}\|_{\mathcal{F}}/\sqrt{2}$ in our implementation.  
The pose calibration is then formulated as an optimization problem:
\begin{align}
 \min_{\Delta R}\quad & \mathcal{L}_{\text{cali}}(\hat{R}, \Delta RR)  \label{eq:canonicalization} \\
 \quad\textrm{s.t.}\quad &\Delta R \in SO(3) 
\end{align}
This problem can be solved by the classical Procrustes analysis~\cite{gower1975generalized}.
In practice, we find that a simple search-based optimization method works reliably.
Concretely, the optimal global rotation $\Delta R^*$ is searched from the set $\{\Delta R_j: \Delta R_j = \hat{R}_jR_j^T\}$.
Moreover, the calibration error $L^*_{\text{cali}}$ is closely related to the noise level of the calibrated pose annotations. 
Large calibration error typically means the failure of calibration and higher level of noise in the calibrated pose annotations (see Sec.~\ref{sec:analysis} for our empirical studies).
Therefore, the calibration error $L^*_{\text{cali}}$ may serve as a heuristic to filter out noisy pose labels. 

\subsection{Absolute pose estimation}\label{sec:regression}
We now could apply any supervised learning methods for pose estimation on our calibrated dataset $\{(x_i, R^{\text{cali}}_i)\}$. 
In this work, we adopt the framework proposed by Xiao~\etal~\cite{xiao2019pose,xiao2021posecontrast} to train our pose estimator. Concretely, we use three Euler angles as our pose representation, including azimuth $\alpha \in [-\pi, \pi]$, elevation $\beta  \in [-\pi/2, \pi/2]$, and roll $\gamma \in [-\pi, \pi]$. 
The Euler angles are decomposed from the rotation matrices $R^{\text{cali}}$ and divided into $Z_\theta$ disjoint angular bins with bin size $B_\theta=\pi / 12$. 
The model is trained to predict the bin indices $y_\theta\in \{1, \dots, Z_\theta\}$ via a classification loss and within-bin offsets $\delta_\theta$ via a regression loss:
\begin{align}
 \mathcal{L}_{\text{abs}} = \sum_{\theta \in {\alpha, \beta, \gamma}} \mathcal{L}_{cls}(y_\theta, p_\theta) + \lambda \mathcal{L}_{\text{reg}}(\delta_\theta, \hat{\delta}_\theta)
\end{align}
where $p_\theta$ is the probability of the object pose in the bin $y_\theta$; 
$\hat{\delta}_\theta \in [0, 1]$ is the predicted offsets within the bin $y_\theta$; 
$(p_\theta, \hat{\delta}_\theta)=g(x)$ are both outputs of our pose estimation model $g(x)$.
We use the cross-entropy loss as the classification loss $\mathcal{L}_{cls}$ and the smooth-L1 loss as the regression loss $\mathcal{L}_{\text{reg}}$; $\lambda$ is the weight on the regression loss ($\lambda=1$ by default).

At the inference time, the pose prediction $\hat{\theta}$ on the input $x$ is obtained via combining the prediction of the bin classifier and the offsets within the predicted angular bin:
\begin{align}
 \hat{\theta} = (j + \hat{\delta}_{\theta,j})B_\theta \quad\text{with}\quad j = \argmax_i p_{\theta,i}
\end{align}
where $p_{\theta,i}$ is the probability of object pose in the $i$-th bin, and $\hat{\delta}_{\theta,j}$ is the predicted offsets within the $i$-th bin.

\section{Experiments}\label{sec:exp}

\paragraph{\textbf{Implementation details.}} 
We use a standard ResNet50 network with three fully-connected layers as our pose estimation model. 
We initialize our model with ImageNet pretrained weights and fine-tune it during training. 
In the first training stage, we do not apply any data augmentation. 
In the second training stage, we use standard data augmentations including in-plane rotation and flipping. 
We conduct hyperparameter search and checkpoint selection on a validation set separate from our training and test set. The validation set consists of 338 non-truncated and non-occluded car images from PASCAL3D+~\cite{xiang2014beyond}.
Similar to prior work~\cite{mustikovela2020self,mariotti2021viewnet,xiao2019pose,xiao2021posecontrast}, we use a tightly cropped object image as the input to our pose estimation model. 
The input image is resized and padded to $224\times 224$.
We use Adam optimizer with learning rate of 1E-4 and weight decay of 5E-4.
In the second training stage we train our model on videos with a calibration error $\mathcal{L}_{\text{cali}}^*$ (Eqn.~\ref{eq:alignerr}) lower than $7^\circ$. 

\paragraph{\textbf{Benchmarks.}} 
We evaluate the performance of our model on the PASCAL3D+ dataset~\cite{xiang2014beyond} which is a  standard benchmark for 3D pose estimation.
The test split in the PASCAL3D+ dataset consists of 308 non-occluded and non-truncated car images collected from PASCAL VOC dataset~\cite{everingham2010pascal}. 
More recently, Mariotti~\etal~\cite{mariotti2021viewnet} reports their results on the ImageNet validation set included in PASCAL3D+ which consisting of 2712 test images of cars.
To make a comparison with Mariotti~\etal, we provide results on both test splits.
Following prior works, we measure the prediction error using the standard geodesic distance $\Delta R = \|\log R_{gt}^TR_{pred}\|_{\mathcal{F}}/\sqrt{2}$ between the estimated rotation matrix $R_{pred}$ and the groundtruth $R_{gt}$. 
We report the median geodesic error (Med.) and the percentage of predictions with error less than $\pi/6$ (Acc.) relative to the groundtruth. 

\paragraph{\textbf{Pose calibration for evaluation.}}
The pose predictions from our model align with human annotations up to a global rotation, due to the difference between the coordinate frame of our model and that of pose annotation tools adopted by the benchmarks. 
To evaluate our model on the benchmarks, similar to prior unsupervised learning methods~\cite{mariotti2021viewnet,mustikovela2020self}, we need to calibrate our pose estimations to the groundtruth annotations. 
Such pose calibration for evaluation is exactly same as our pose calibration step described in Sec~\ref{sec:cano}.
Specifically, we estimate a global calibration matrix $\Delta R$ such that $\Delta R R_{pred}$ equals the human annotations $R_{gt}$. 
We formulate the pose calibration as an optimization problem and solve it via a simple search-based method (see more details in Sec~\ref{sec:cano}).
The calibration matrix $\Delta R$ is obtained via solving the optimization problem on 100 car images randomly sampled from the training set of PASCAL3D+.

\subsection{Pose estimation}

\paragraph{\textbf{Quantitative results.}}  Tab.~\ref{tab:score} provides quantitative comparisons with prior unsupervised pose estimation works on PASCAL3D+ test set. 
Our method significantly outperforms the existing SfM-based methods~\cite{sedaghat2015unsupervised,novotny2017learning}. 
Similar to ours, these models are trained on video data with pose annotations from SfM.
However, they rely on SfM with SIFT~\cite{lowe2004distinctive} and nearest neighbor (NN) matching, which fails to provide high-quality pose estimations (see more details in Sec~\ref{sec:analysis}).
For this reason, prior SfM-based models collect videos by slowly moving a camera around \emph{static} cars to avoid large motion blur. 
This tedious procedure limits the size of existing car video datasets.
For example, the FreiburgCars dataset~\cite{sedaghat2015unsupervised} consists of 52 car videos; the EPFL car dataset~\cite{ozuysal2009pose} provides only 20 car videos. 
In comparison, our video dataset (consisting of 313 videos) is easy to collect and prevalent on the Internet. 
SfM, coupled with the recent progress in object detection~\cite{he2017mask} and feature matching~\cite{sarlin2020superglue}, provides robust and accurate pose estimations on our in-the-wild videos, which is the key to success of our framework.
Our model trained on the accidental turntable dataset achieves higher pose prediction accuracy than when trained on the FreiburgCars dataset.

In comparison with analysis-by-synthesis frameworks~\cite{mustikovela2020self,mariotti2021viewnet}, our prediction accuracy is significantly higher than that of SSV model~\cite{mustikovela2020self} which is trained on the CompCars dataset~\cite{yang2015large} (consisting of 137,000 real car images). 
ViewNet~\cite{mariotti2021viewnet} achieves the highest performance on PASCAL3D+ among existing unsupervised learning methods.
However, this method relies on 3D models from ShapeNet~\cite{chang2015shapenet} to generate a highly curated dataset with controlled variations in viewpoint, translation, lighting, and background, etc. 
Moreover, ViewNet has a harder time learning from real videos (\eg FreiburgCars~\cite{sedaghat2015unsupervised}) where its performance drops remarkably.
We're unable to train ViewNet on our own dataset, as the source code has not yet been released.

\begin{table*}
\begin{center}
\caption{\label{tab:score} \textbf{Pose estimation on PASCAL3D+ test sets.} 
We make comparisons with supervised learning methods trained with human annotations (dubbed Anno.) and unsupervised pose estimation models based on Structure-from-Motion (dubbed SfM) or Analysis-by-Synthesis (dubbed AbS).
$^\ast$ViewNet ignores the in-plane rotation in the evaluation and reports the results on ImageNet validation set.
}
\begin{tabular}{llcc cc c }
\hline\noalign{\smallskip}
  & Methods & Supervision & Trainset & Testset & Acc.($\%$) $\uparrow$ & Med.($\degree$) $\downarrow$ \\
\noalign{\smallskip}
\hline
\noalign{\smallskip}
\parbox[t]{2mm}{\multirow{4}{*}{\rotatebox[origin=c]{90}{Super.}}} 
 &Tulsiani~\etal~\cite{tulsiani2015viewpoints} & Anno. & PASCAL3D+ & VOC &89  & 9.1\\
 &Mahendran~\etal~\cite{mahendran20173d} & Anno.  & PASCAL3D+  & VOC & -- & 8.1 \\
 &Liao~\etal~\cite{liao2019spherical} & Anno.  & PASCAL3D+ & VOC & 93 & 5.2 \\
 &Grabner~\etal~\cite{grabner20183d} & Anno.   & PASCAL3D+ & VOC & 94 & 5.1 \\
\hline
\parbox[t]{2mm}{\multirow{7}{*}{\rotatebox[origin=c]{90}{Unsupervised}}} 
 &VPNet~\cite{sedaghat2015unsupervised} & SfM  & FreiburgCars & VOC & -- &   49.6 \\
 &VpDRNet~\cite{novotny2017learning} & SfM  & FreiburgCars & VOC & -- &  29.6 \\
 &SSV~\cite{mustikovela2020self} & AbS  & CompCars & VOC & 67 & 10.1 \\ 
 &Ours & SfM & FreiburgCars & VOC & 72 & 15.7 \\
 &Ours & SfM & Acci.Turn. & VOC & 75 & 15.8 \\
 \cline{2-7}
 &ViewNet$^\ast$~\cite{mariotti2021viewnet} & AbS & ShapeNet & ImageNet  & 88 & 5.6 \\
 &ViewNet$^\ast$~\cite{mariotti2021viewnet} & AbS & FreiburgCars & ImageNet  &  61 & 16.1 \\
  &Ours & SfM & FreiburgCars & ImageNet & 84 & 15.0 \\
 &Ours & SfM & Acci.Turn. & ImageNet & 86 & 14.8 \\
\hline
\end{tabular}
\end{center}
\end{table*}

\begin{figure}
\centering
\includegraphics[width=\linewidth]{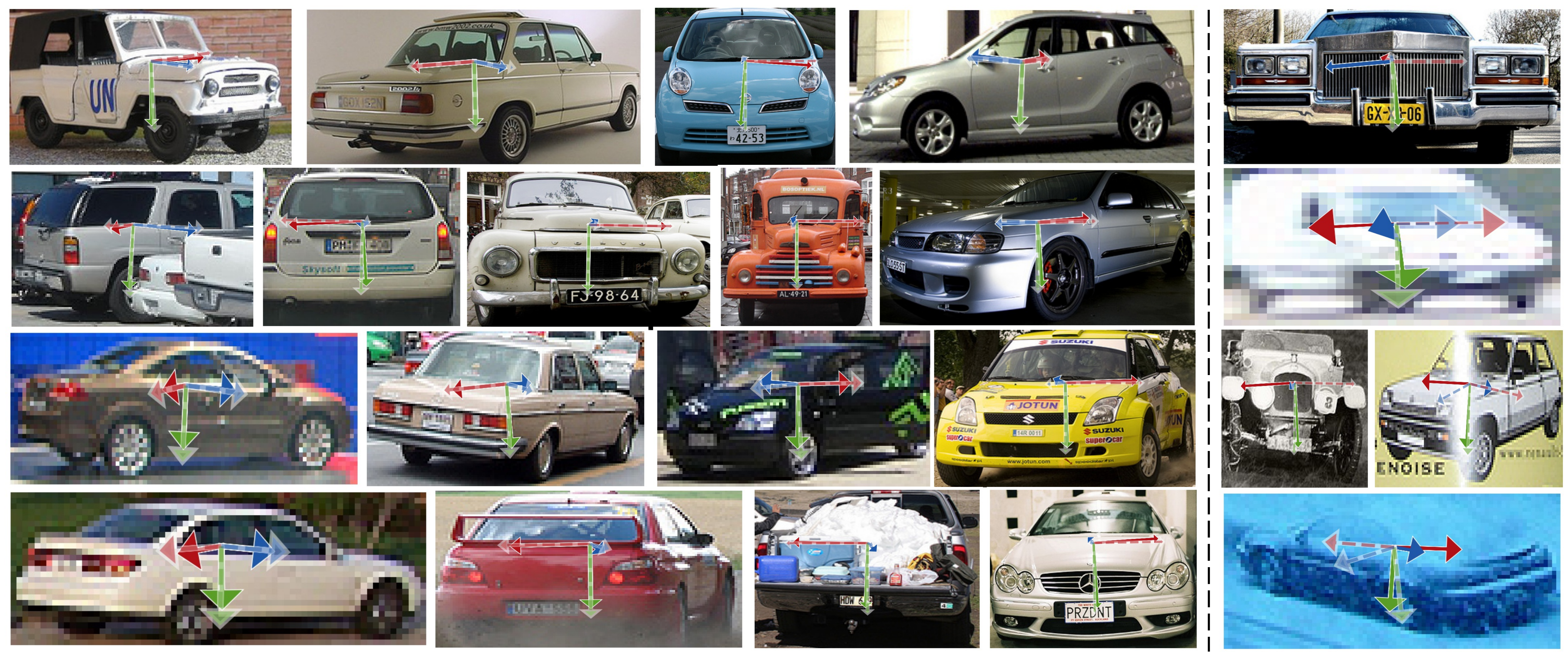}
\caption{\textbf{Pose prediction on Pascal3D+ test set.} \textbf{Left}: our model achieves high accuracy of pose estimation on cars in diverse appearance, poses, and shapes. \textbf{Right}: the performance drops on large, occluded objects (1st row), low-resolution images (2nd row) or out-of-domain data (last two rows). The solid arrows indicate the pose predictions from our model and the dashed arrows are the groundtruth annotations. The \textcolor{blue}{blue} arrow directs towards the frontal side of cars and the \textcolor{red}{red} points toward the right side. The angular distances between the predictions and the groundtruth are less than $7^\circ$ for examples on the left while higher than $90^\circ$ on the failure cases.}
\label{fig:prediction}
\end{figure}

\paragraph{\textbf{Qualitative results.}} 
Fig.~\ref{fig:prediction} visualizes our pose predictions on Pascal3D+ test set. Our model provides accurate pose estimation on diverse cars in terms of appearance, poses and shapes. 
The performance of our model drops in several cases: the object is highly occluded; the image is in low resolution; the domain gap between the input and our dataset is large (\eg cartoon cars, snow-covered cars). 
These issues can be potentially relieved by collecting more videos to further enrich the diversity of cars in our dataset.

\subsection{Analysis}\label{sec:analysis}

\begin{figure}
\centering
\includegraphics[width=\linewidth]{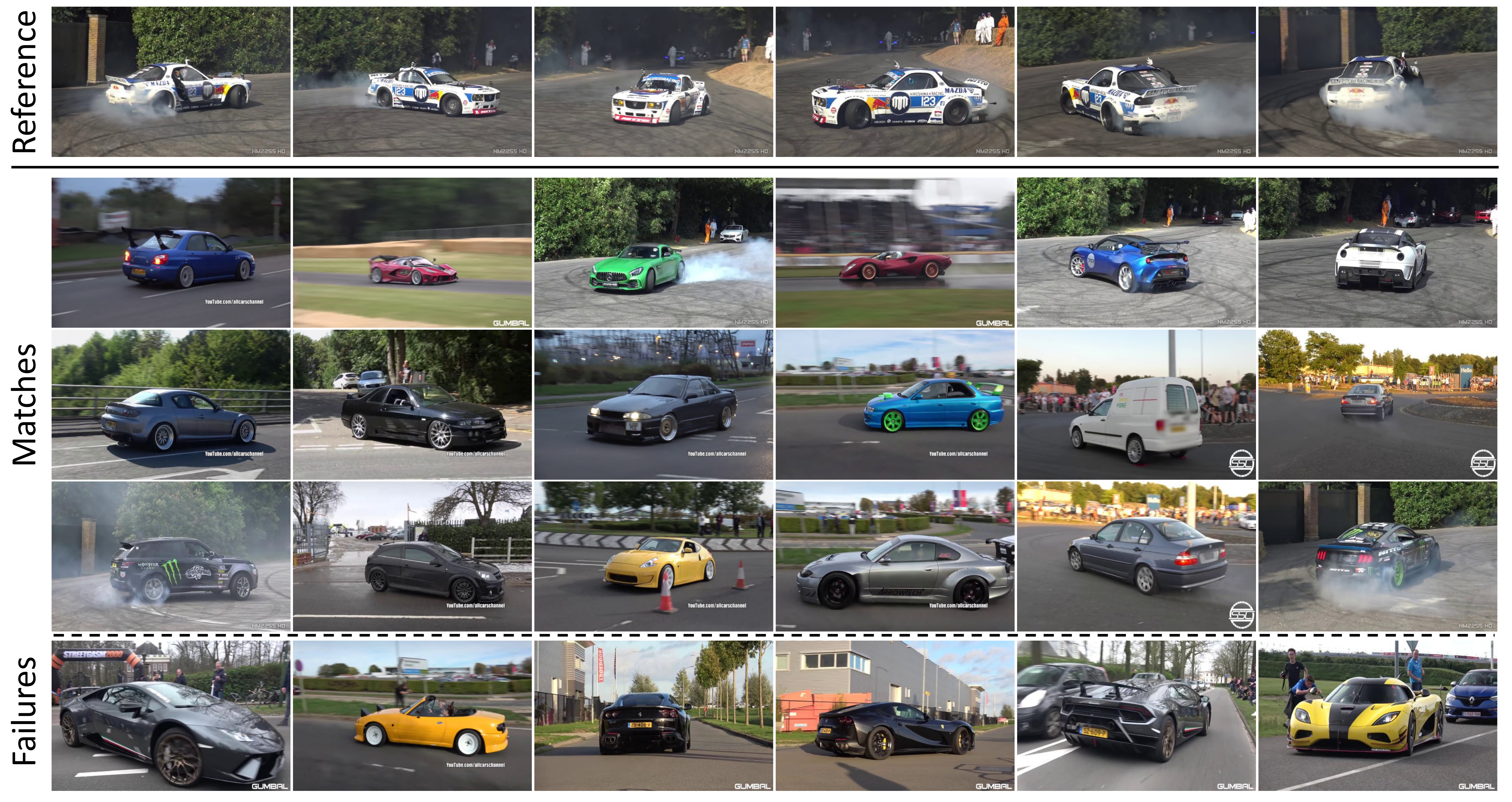}
\caption{\textbf{Canonical pose emerges in our first training stage (Sec.~\ref{sec:relative})}. For each reference images (top), we present four matches (including one failure case) of which the pose annotations have less than $5^\circ$ angular distance to that of the reference frame. 
The calibration error $\mathcal{L}_{\text{cali}}^*$ (Eqn.~\ref{eq:alignerr}) is higher than $25^\circ$ on these failure cases while lower than $10^\circ$ on the well-calibrated video instances. This provides us a heuristic to filter out noisy annotations.}
\label{fig:cano}
\end{figure}

\paragraph{\textbf{The emergence of canonical pose.}} The key to the success of the proposed model is the
emergence of canonical pose in our first training stage.
Fig.~\ref{fig:cano} provides images from our dataset with similar pose annotations after the calibration step (Sec.~\ref{sec:cano}). 
On the one hand, Fig.~\ref{fig:cano} clearly demonstrates that the calibrated pose annotations well align in a uniform frame.
On the other hand, the calibration fails on several videos due to the limited performance of our stage-one model (Fig.~\ref{fig:cano} bottom).
A typical failure case is that the pose predictor misidentifies the frontal view of a car as the rear view. 
Such failure cases of pose calibration introduce noisy pose annotations into our dataset.
Fortunately, we find that the noise level of the annotations are closely correlated with the calibration error $\mathcal{L}_{\text{cali}}^*$ (Eqn.~\ref{eq:alignerr}).
We thus use the calibration error $\mathcal{L}_{\text{cali}}^*$ as a heuristic to filter out noisy annotations in our second training stage. 

\begin{figure}
    \newcommand\wifig{0.45\linewidth}
    \centering
     \begin{tabular}{cc}
    \includegraphics[width=\wifig]{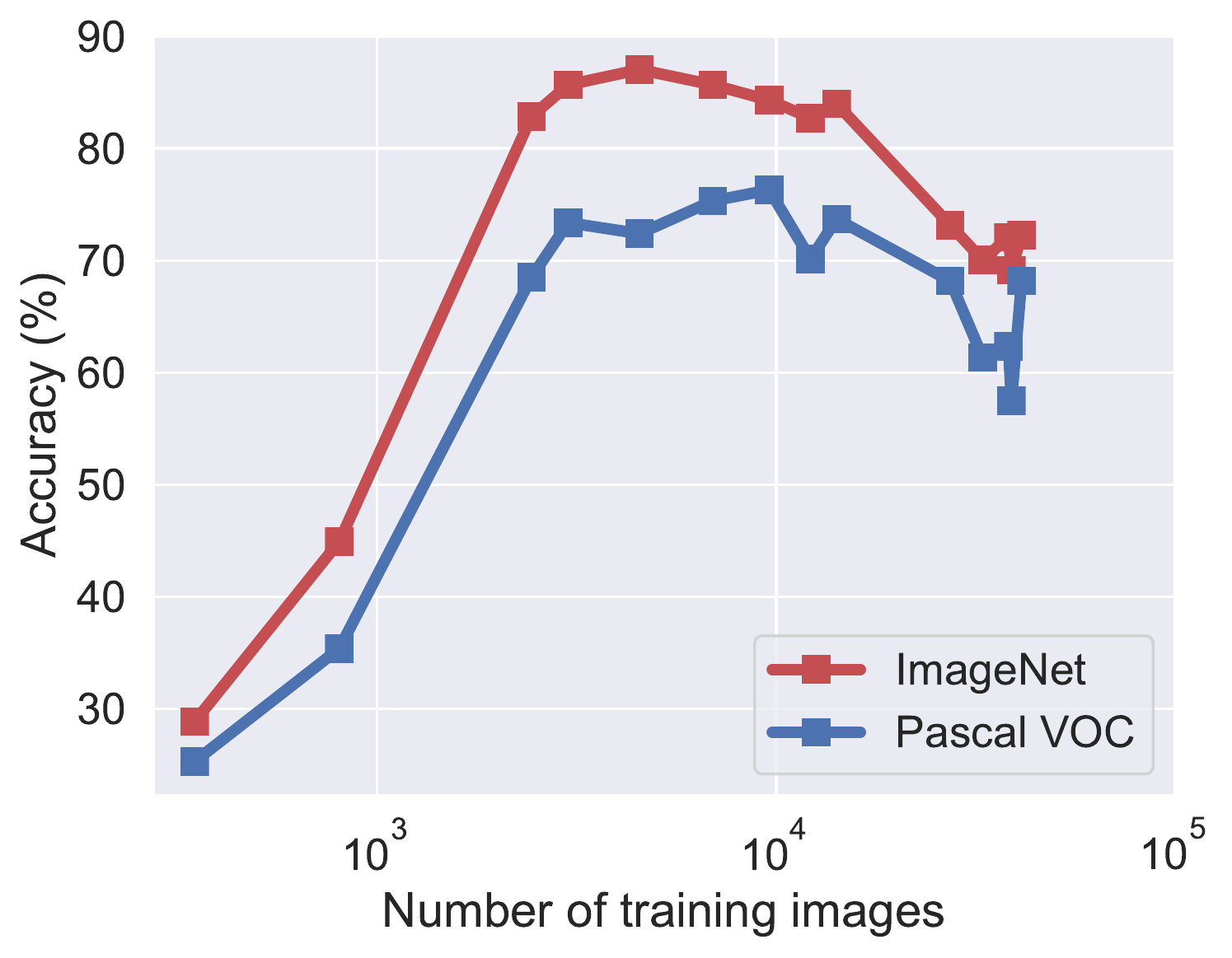} &
    \includegraphics[width=\wifig]{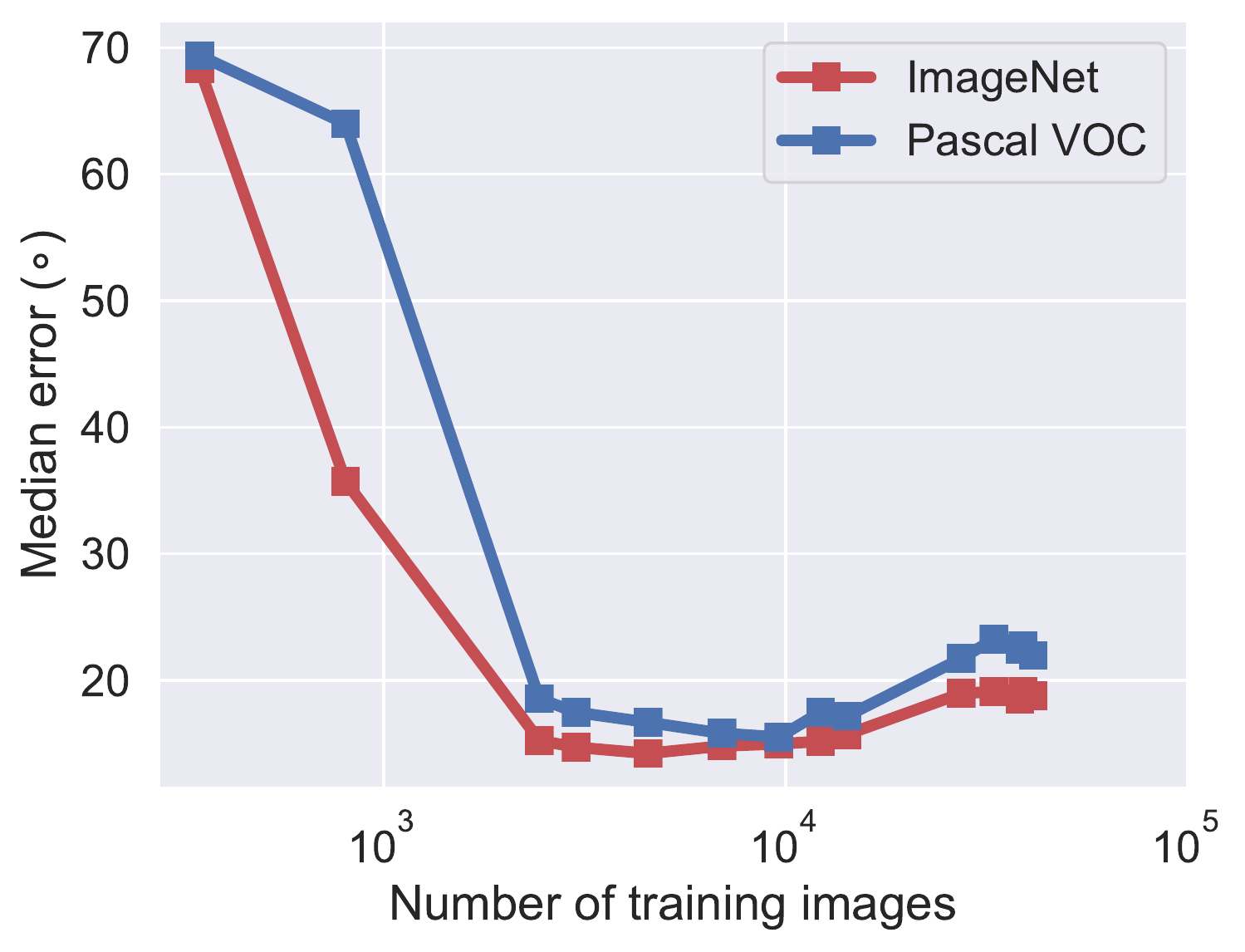}\\
    \end{tabular}
    \caption{\textbf{The effect of annotation noise level on 3D pose prediction.} 
    We report the performance of our pose estimation model under different noise level of pose annotations.
    Higher level of annotation noise corresponds to larger number of training images. 
    We report both prediction accuracy (left panel) and median error (right panel) on two test splits included in PASCAL3D+. 
    }
    \label{fig:noise-level}
\end{figure}

\paragraph{\textbf{The effect of noise level in the annotations.}} 
We use the calibration error $\mathcal{L}_{\text{cali}}^*$ (Eqn.~\ref{eq:alignerr}) as an indicator of the noise level of the pose annotations. 
Higher threshold on the calibration error corresponds to larger number of training images yet more noisy annotations, and vice versa.
Fig.~\ref{fig:noise-level} presents the performance of our model under different noise levels of the annotations.
It demonstrates that neither clean-yet-small data nor large-yet-noisy data leads to higher performance than mid-size data with mid-level noise. 
We provide more analysis in Appendix~\ref{sec:more-analysis}.

\paragraph{\textbf{The effect of two-stage training.}} 
As demonstrated in Fig~\ref{fig:cano}, the model trained in the first stage provides a tool to calibrate the pose annotations of our dataset. 
However, the  performance of the stage-one model lags behind the state-of-the-art analysis-by-synthesis frameworks (\eg SSV~\cite{mustikovela2020self} and ViewNet~\cite{mariotti2021viewnet}). 
We hypothesize that training to predict the relative pose is a suboptimal learning strategy for the task of absolute pose estimation. 
As shown in Tab.~\ref{tab:two-stage}, the model trained in our second training stage significantly outperforms the the one trained in the first stage.
This suggests that learning with the absolute pose annotations is indeed a more effective training method.
However, our stage-two training is not possible without the pose calibration and stage-one model. 
Therefore, the proposed two training stages are complementary and both play an important role in our framework.

\setlength{\tabcolsep}{4pt}
\begin{table}
\begin{center}
\caption{\label{tab:two-stage} \textbf{The effect of two-stage training on 3D pose prediction.}
The second stage trains the model to regress absolute pose after using the first stage model to
calibrate the relative pose annotations. This procedure leads to a significant improvement in
pose estimation accuracy ($\%$) and median error ($\degree$), in spite of the training datasets. }
\label{tab:two-stage}
\begin{tabular}{cc cc cc }
\hline\noalign{\smallskip}
\multirow{2}{*}{Trainset} & \multirow{2}{*}{Stage} & \multicolumn{2}{c}{PASCAL VOC} & \multicolumn{2}{c}{ImageNet} \\
  & & Acc. $\uparrow$ & Med. $\downarrow$ & Acc. $\uparrow$ & Med. $\downarrow$ \\
\noalign{\smallskip}
\hline
\multirow{2}{*}{Acci. Turn.} & 1 & 42 & 38.8 & 46 & 32.9 \\
& 2 & \textbf{75} & \textbf{15.8} & \textbf{86} & \textbf{14.8} \\
\hline
\multirow{2}{*}{FreiburgCars} & 1 & 36 & 44 & 47 & 31.9 \\
& 2 & \textbf{72} & \textbf{15.7} & \textbf{84} & \textbf{15.0} \\
\hline
\end{tabular}
\end{center}
\end{table}

\paragraph{\textbf{The effect of network initialization.}} The recent self-supervised learning (SSL)~\cite{he2020momentum,chen2020improved} has significantly improves the unsupervised pose estimation~\cite{cheng2021equivariant} and part discovery~\cite{saha2021ganorcon}. 
We initialize our pose estimation network with ImageNet-pretrained models by default. 
However, ImageNet classification labels require extensive human labors. 
A natural question is how the recent SSL methods help us further reduce the requirement of human annotations. 
Tab.~\ref{tab:init} provides a comparison of different initialization strategies.
Supervised ImageNet pretraining and unsupervised contrastive pretraining~\cite{he2020momentum,chen2020improved} have similar performance in the task of pose estimation, while both outperforms the random initialization in a large margin. 

\setlength{\tabcolsep}{5pt}
\begin{table}
\begin{center}
\caption{\label{tab:initialization} \textbf{The effect of network initialization on 3D pose prediction.} 
ImageNet pretrained models provide significant improvement over random initialized ones but self-supervised counterparts are competitive alternatives without having to resort to extra human annotations.
}
\label{tab:init}
\begin{tabular}{lc cc cc }
\hline\noalign{\smallskip}
\multirow{2}{*}{Initialization} & \multicolumn{2}{c}{PASCAL VOC} & \multicolumn{2}{c}{ImageNet} \\
  & Acc. $\uparrow$ & Med. $\downarrow$ & Acc. $\uparrow$ & Med. $\downarrow$ \\
\noalign{\smallskip}
\hline
Random & 58 & 25 & 70 & 20.2 \\
Contrastive~\cite{he2020momentum} & 74 & 15.7 & 85 & \textbf{14.3}\\
ImageNet & \textbf{75} & \textbf{15.8} & \textbf{86} &  14.8 \\
\hline
\end{tabular}
\end{center}
\end{table}

\begin{figure}[ht!]
\centering
\includegraphics[width=\linewidth]{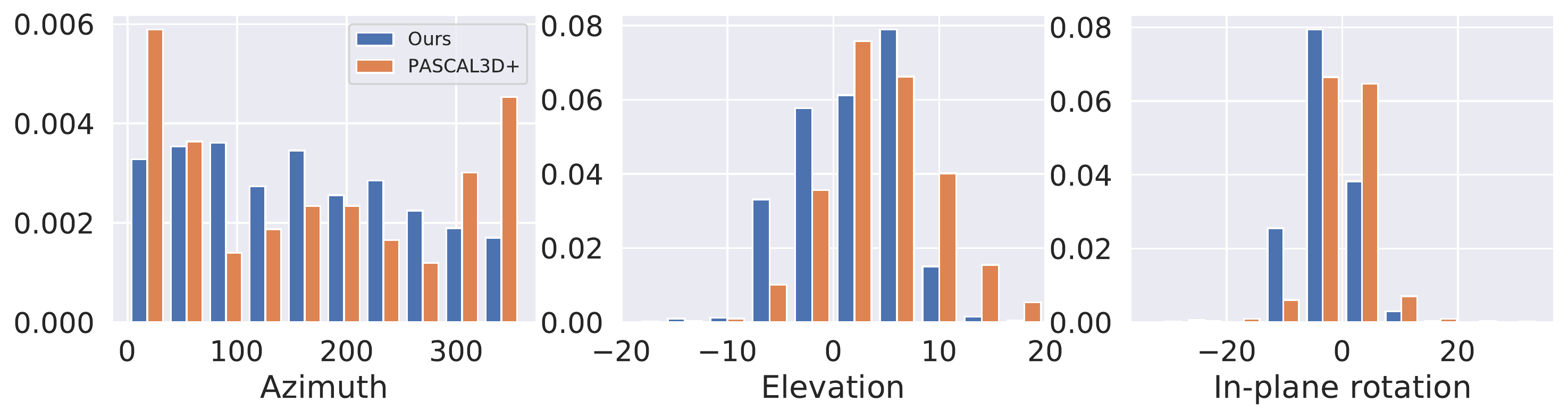}
\captionof{figure}{Distribution of the poses in the proposed accidental turntable dataset and the PASCAL3D+.}
\label{fig:pose-dist}
\end{figure}

\paragraph{\textbf{Pose distribution.}}
Figure~\ref{fig:pose-dist} compares the pose distribution of the Acci-
dental Turntables dataset and PASCAL3D+. The distribution of azimuth is more balanced in our dataset, where PASCAL3D+ has more cars with large elevations.

\begin{figure}[ht!]
\centering
\includegraphics[width=0.9\linewidth]{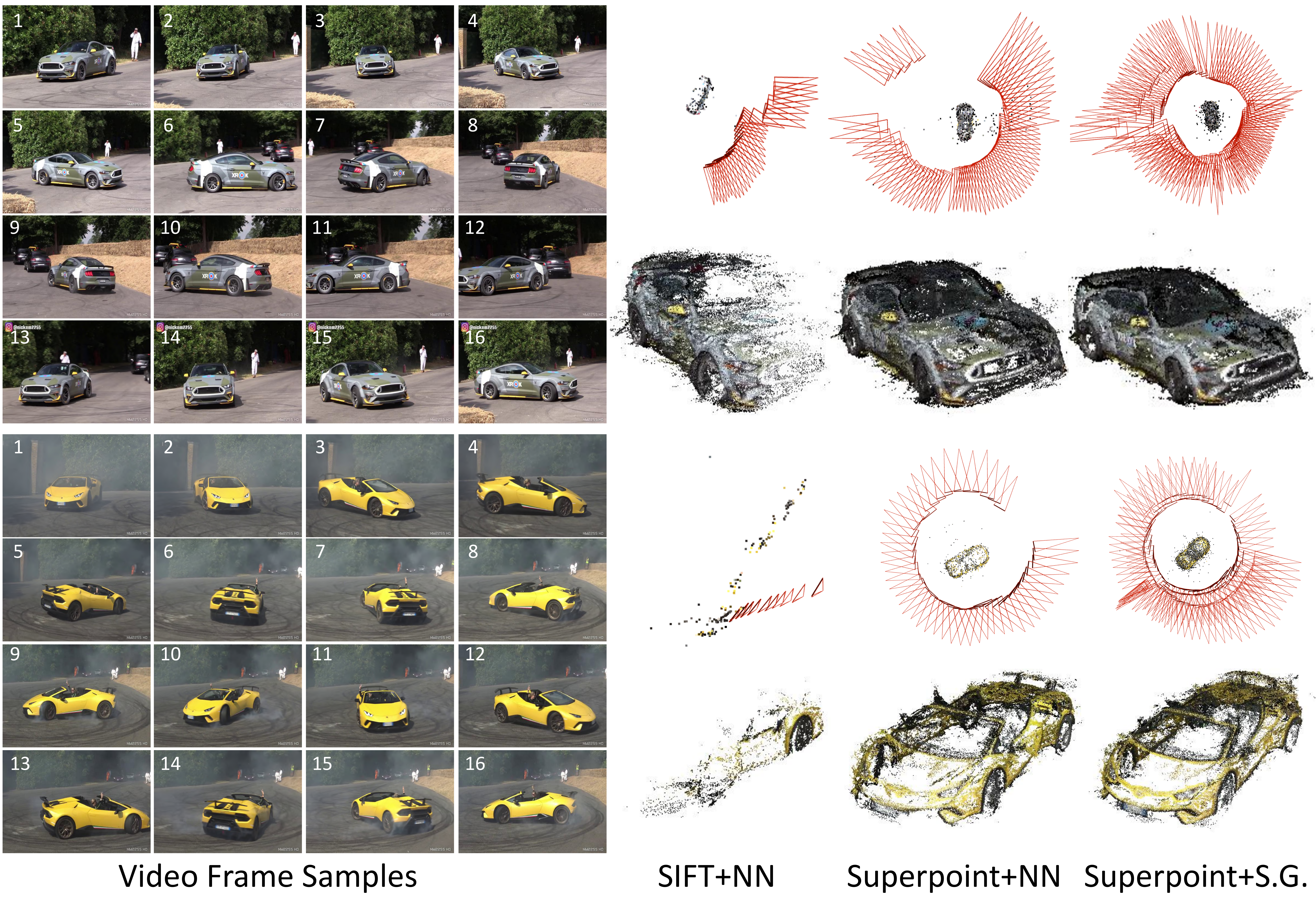}
\caption{\textbf{Feature extraction and matching for structure-from-motion.} 
\textbf{Left}: video samples from the proposed accidental turntable dataset.  
\textbf{Right}: pose estimations (top) and dense 3D reconstruction (bottom) under different feature extraction (SIFT~\cite{lowe2004distinctive} or Superpoint~\cite{detone2018superpoint}) and matching (nearest neighbor (NN) or SuperGlue (S.G.)~\cite{sarlin2020superglue}) algorithms.
The \textcolor{red}{red} square pyramids indicate the location of the estimated camera pose. 
Each video consists of more than 200 frames and the car turns around $720^{\circ}$. 
}
\label{fig:SfM}
\end{figure}

\paragraph{\textbf{Feature extraction and matching for SfM.}}
Feature extraction and matching is the core of SfM algorithms.
The classical SIFT~\cite{lowe2004distinctive} and simple nearest neighbor matching (NN) remains the default components in popular SfM packages (\eg COLMAP~\cite{schonberger2016structure,schonberger2016pixelwise}), despite of the recent success of learning-based methods~\cite{detone2018superpoint,sarlin2020superglue}.
We observe that SfM with SIFT and NN does not work reliably on our in-the-wild video dataset.
Fig.~\ref{fig:SfM} compares the 3D reconstruction and pose estimation from COLMAP under different feature extraction and matching algorithms on two videos from our dataset.
SfM with SIFT and NN only provides partial 3D reconstruction and pose estimation on a small subset of frames. 
Its performance drops significantly on texture-free objects (Fig.~\ref{fig:SfM} bottom).
Simply replacing SIFT with Superpoint~\cite{detone2018superpoint} leads to more complete 3D reconstruction and pose estimations.
SfM with Superpoint and SuperGlue~\cite{sarlin2020superglue} provides the highest quality of shape reconstruction and pose estimations.
Our experimental results can be explained by the following observations:
SIFT detects few interest points on most cars due to the texture-free surface;
SIFT extracts features in a small local region, which results in large ambiguity in matching duplicated patterns (\eg frontal and rear wheels of a car);
large motion blur further destabilizes the feature matching process; 
In comparison, Superpoint provides rich interest points even on texture-free regions; 
Lastly, SuperGLUE aggregates long-range contextual information via attention mechanism, 
which we find significantly reduces the ambiguity in matching repeated patterns. 
Fig.~\ref{fig:sfm-more} provides more examples from our accidental turntable dataset. 
The performance of SfM may drop on highly-occluded objects (\eg the car is occluded by smoke in  Fig.~\ref{fig:sfm-more} bottom).

\begin{figure}
\centering
\includegraphics[width=0.9\linewidth]{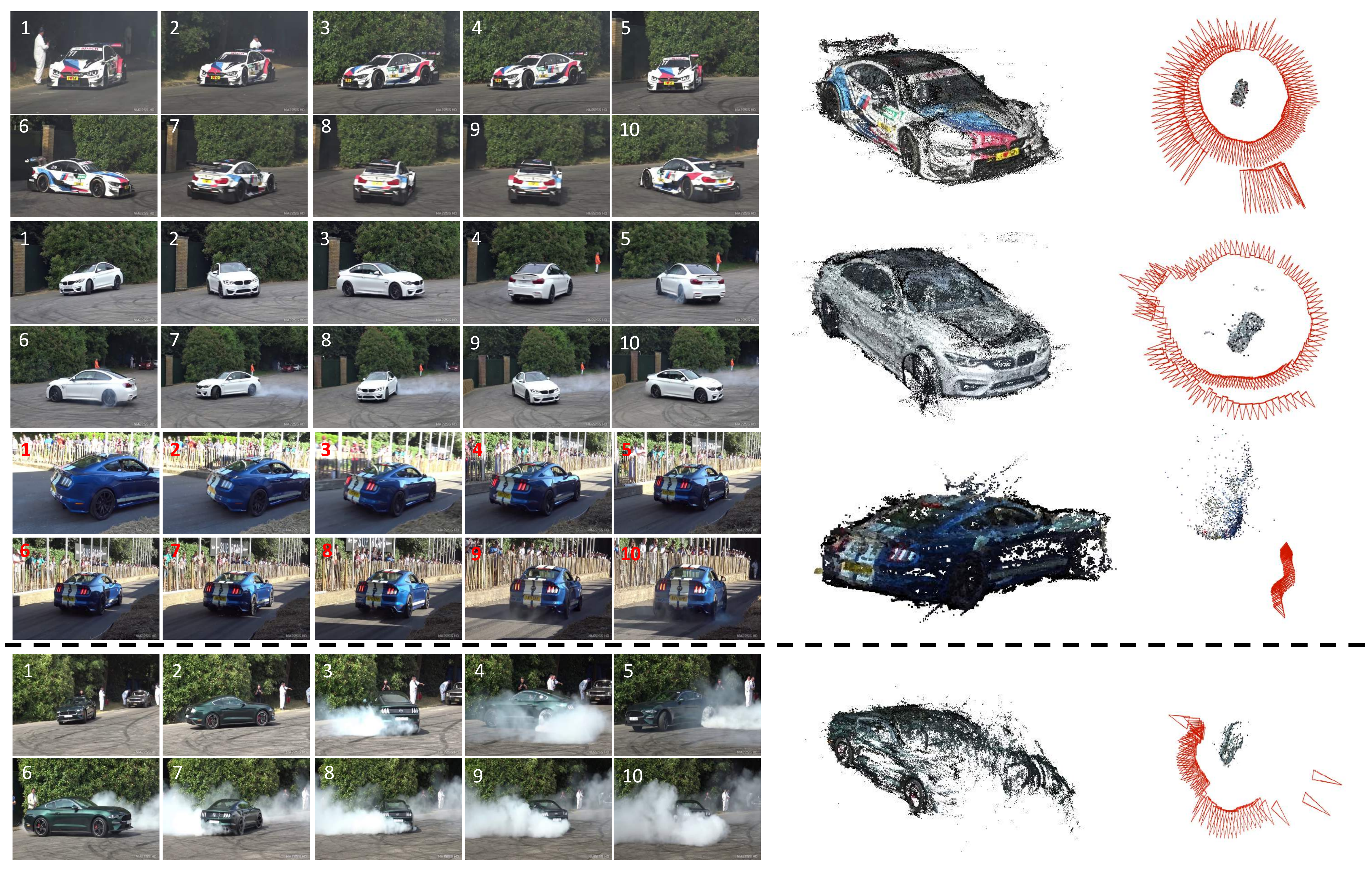} \\
\captionof{figure}{\textbf{More examples from the Accidental Turntables dataset.} SfM provides accurate 3D reconstructions and pose estimations on either texture-rich (\textbf{1st row}) or texture-free (\textbf{2nd row}) objects, as well as objects moving along a straight line without any turns (\textbf{3rd row}). The performance drops on highly-occluded objects (\textbf{bottom}). See more examples in Appendix~\ref{sec:more-dataset}.}
\label{fig:sfm-more}
\end{figure}

\begin{figure}
\centering
\includegraphics[width=\linewidth]{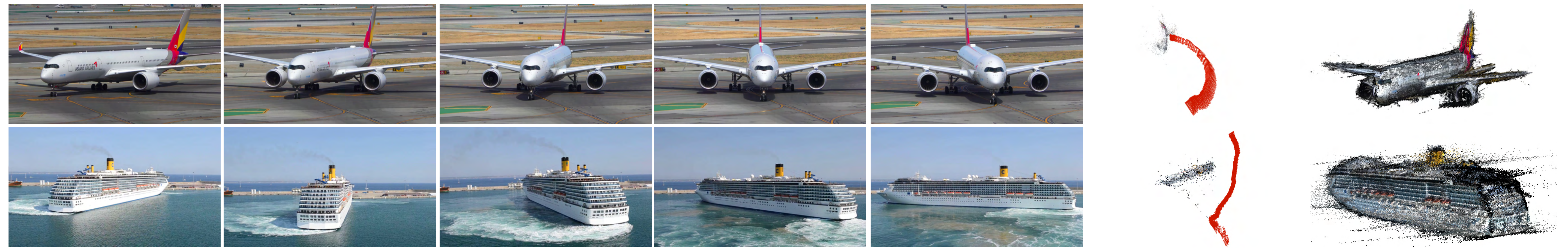} \\
\captionof{figure}{\label{fig:airplane} Accidental turntables for airplanes and cruise. \textbf{Left:} video frame samples. \textbf{Right:} pose estimation and 3D reconstruction from structure-from-motion. }
\label{fig:other-categories}
\vspace{-2mm}
\end{figure}

\paragraph{\textbf{Extension to other categories.}}
There are a fair number of turntable videos for other categories on Youtube. For example, airplanes turn along the runway 
(\eg, \href{https://youtu.be/5EgzWcI-Io0?t=465}{video1}, \href{https://youtu.be/5EgzWcI-Io0?t=1253}{video2}, \href{https://youtu.be/khesztRJKUw?t=933}{video3},  \href{https://youtu.be/khesztRJKUw?t=1086}{video4}); 
landing or takeoff of airplanes usually induces more than 90-degree pose changes relative to the camera 
(\eg, \href{https://youtu.be/Z7CutgNEMfA?t=30}{video5}, \href{https://youtu.be/9qHLKmuxFf8?t=44}{video6});
cruises turn (\eg \href{https://www.youtube.com/watch?v=CgaJgRdI3FQ&t=3s}{video7}).
Fig.~\ref{fig:other-categories} shows SfM with Superpoint and SuperGlue provides reasonable pose estimation and 3D reconstruction on these categories. 
Even though we focus on cars in this work, our dataset is much larger, easier to collect, and more useful to train a pose estimator than existing car datasets (\eg, FreiburgCars). 

\section{Discussion and Conclusion}

We propose to learn 3D pose estimation models from a new source of data: videos where objects turn.
We demonstrate that classical structure-from-motion algorithms, coupled with the recent advances in feature matching and object detection, provide surprisingly accurate pose estimations and 3D reconstructions on in-the-wild car videos. 
We also provide a novel learning framework which successfully trains a high-quality 3D pose predictor on the collected video datasets.
We plan to release our \textbf{Accidental Turntable dataset} along with the pose estimations and 3D reconstructions from the enhanced SfM for the research community.

\clearpage
\appendix
\noindent{\Large \textbf{Appendix}}
 
\section{Accidental Turntables Dataset}\label{sec:more-dataset}

\paragraph{Data source}
We use 6 Youtube videos as the source of our Accidental Turntables dataset including
\href{https://www.youtube.com/watch?v=8rFNRri8-TI}{video1},
\href{https://www.youtube.com/watch?v=Aqacvi267kk&t=53s}{video2},
\href{https://www.youtube.com/watch?v=5YbJhq1nltw}{video3},
\href{https://www.youtube.com/watch?v=Wl_8mypokYo}{video4},
\href{https://www.youtube.com/watch?v=hQWJC28WYNc}{video5},
\href{https://www.youtube.com/watch?v=MB7lzwzZOio}{video6}.

\paragraph{More data samples and 3D pose annotations}
Fig.~\ref{fig:sfm-more} provides more examples from our Accidental Turntables dataset. 

\begin{figure}[ht!]
\centering
\includegraphics[width=\linewidth]{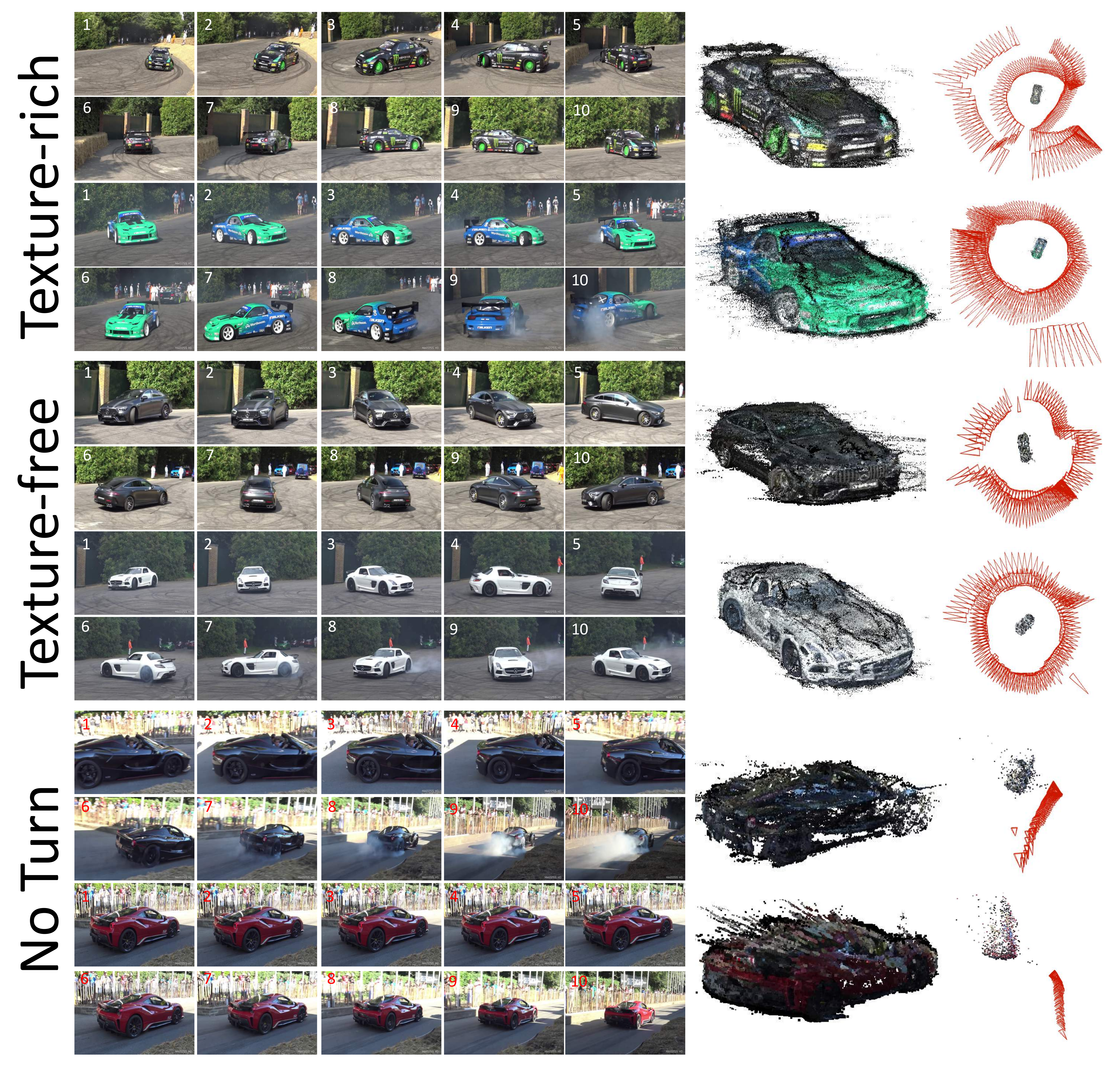}
\vspace{-3mm}
\caption{\textbf{Samples from the Accidental Turntables dataset.} SfM provides accurate 3D reconstructions (\textbf{middle}) and pose estimations (\textbf{right}) on either texture-rich (\textbf{1st row}) or texture-free (\textbf{2nd row}) objects, as well as objects moving along a straight line without any turns (\textbf{3rd row}).}
\label{fig:sfm-more}
\end{figure}

\section{More Analysis}\label{sec:more-analysis}

\begin{figure}[ht!]
    \newcommand\wifig{0.45\linewidth}
    \centering
     \begin{tabular}{cc}
    \includegraphics[width=\wifig]{figs/acc.pdf} & 
    \includegraphics[width=\wifig]{figs/med.pdf} \\
    \multicolumn{2}{c}{\textbf{ImageNet pretraining}} \\
    \includegraphics[width=\wifig]{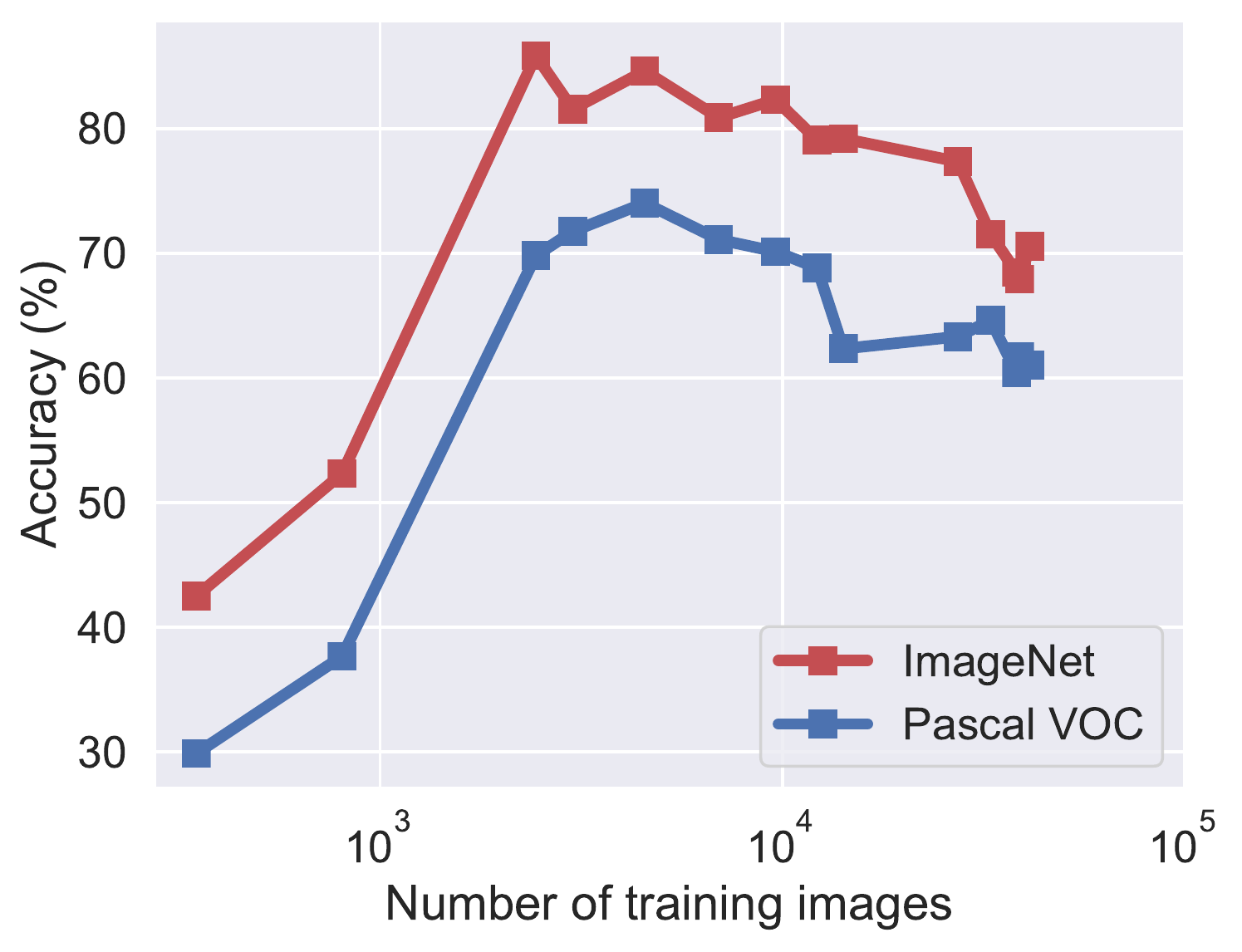} &
    \includegraphics[width=\wifig]{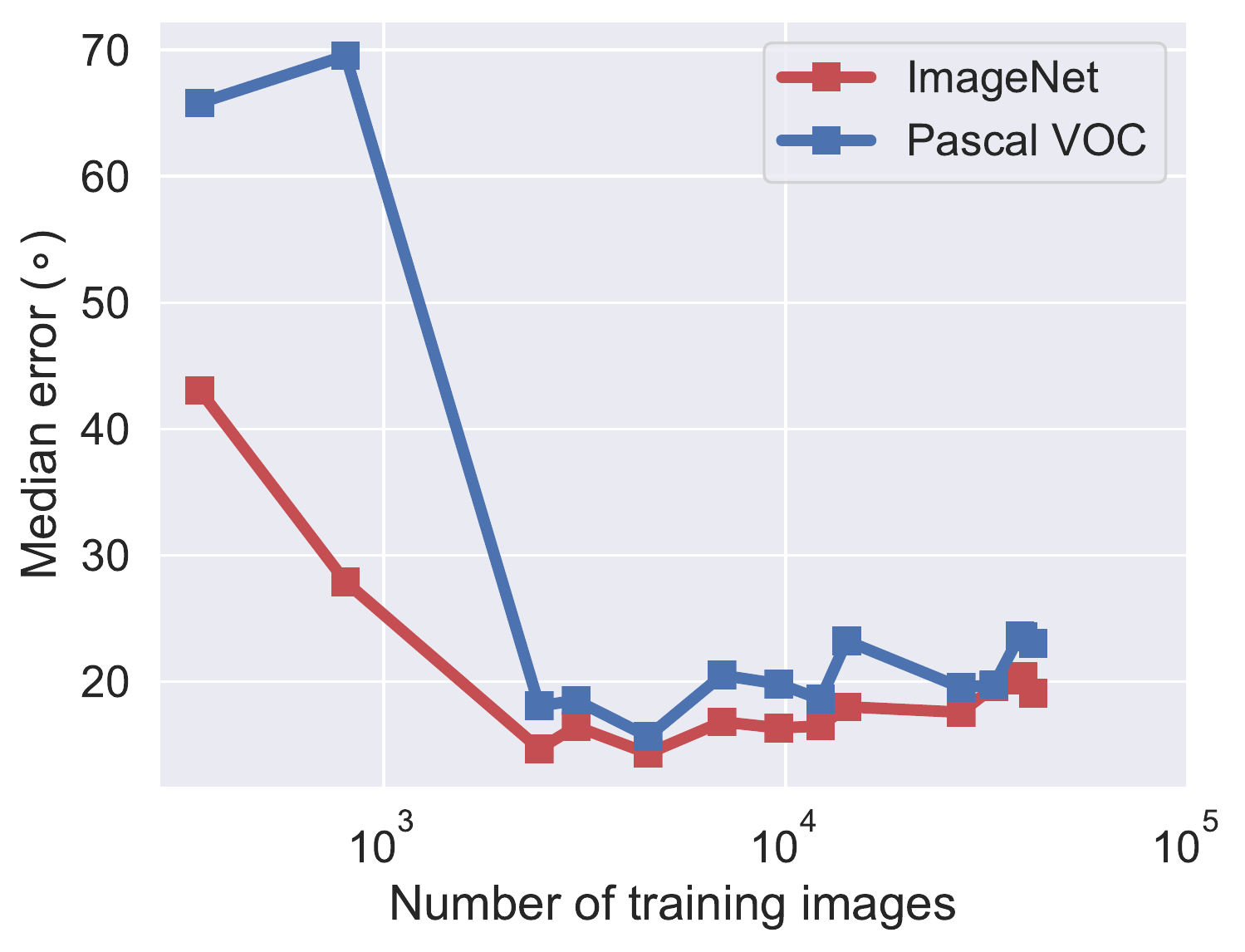} \\
    \multicolumn{2}{c}{\textbf{Contrastive pretraining~\cite{chen2020improved}}} \\
    \includegraphics[width=\wifig]{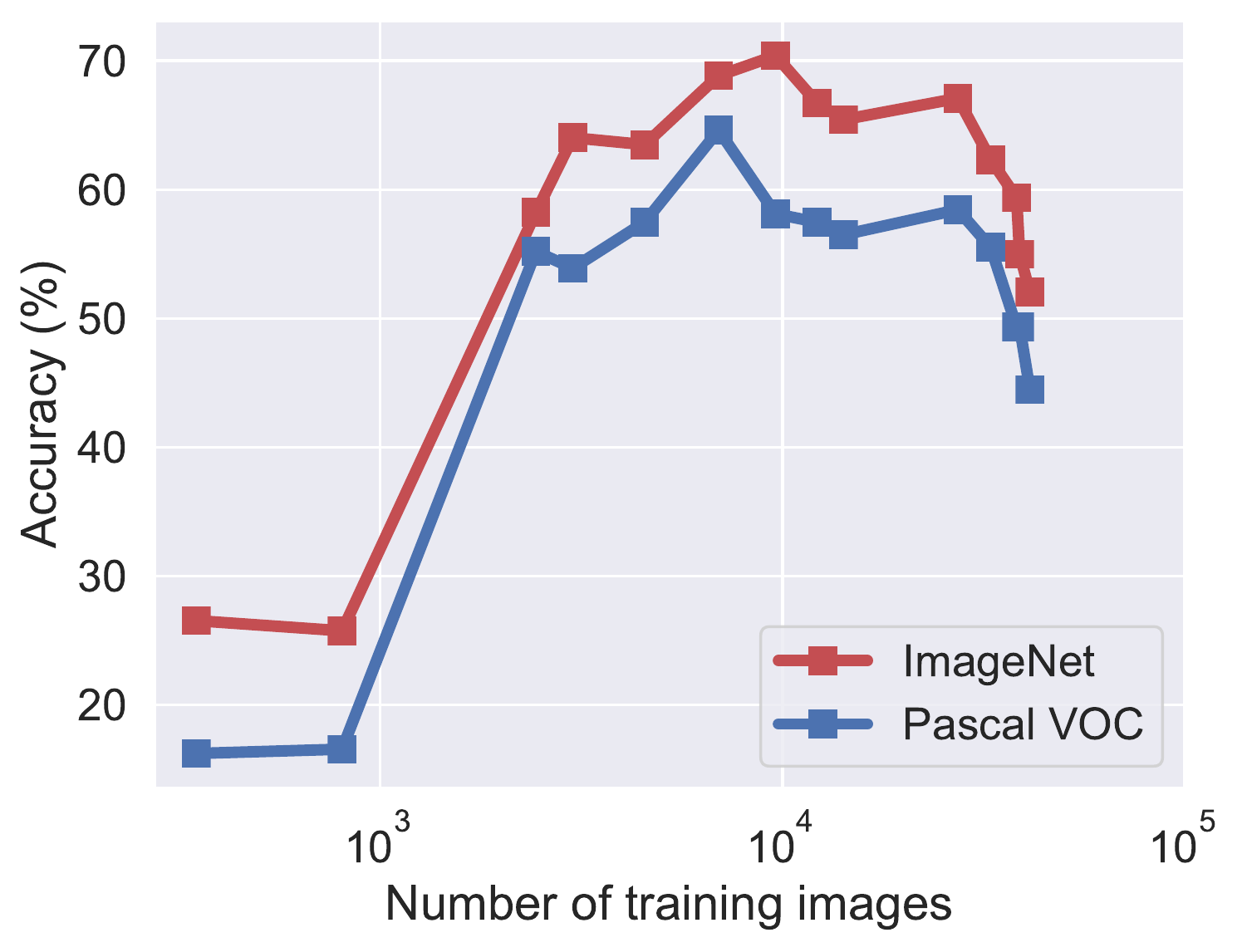}  &
    \includegraphics[width=\wifig]{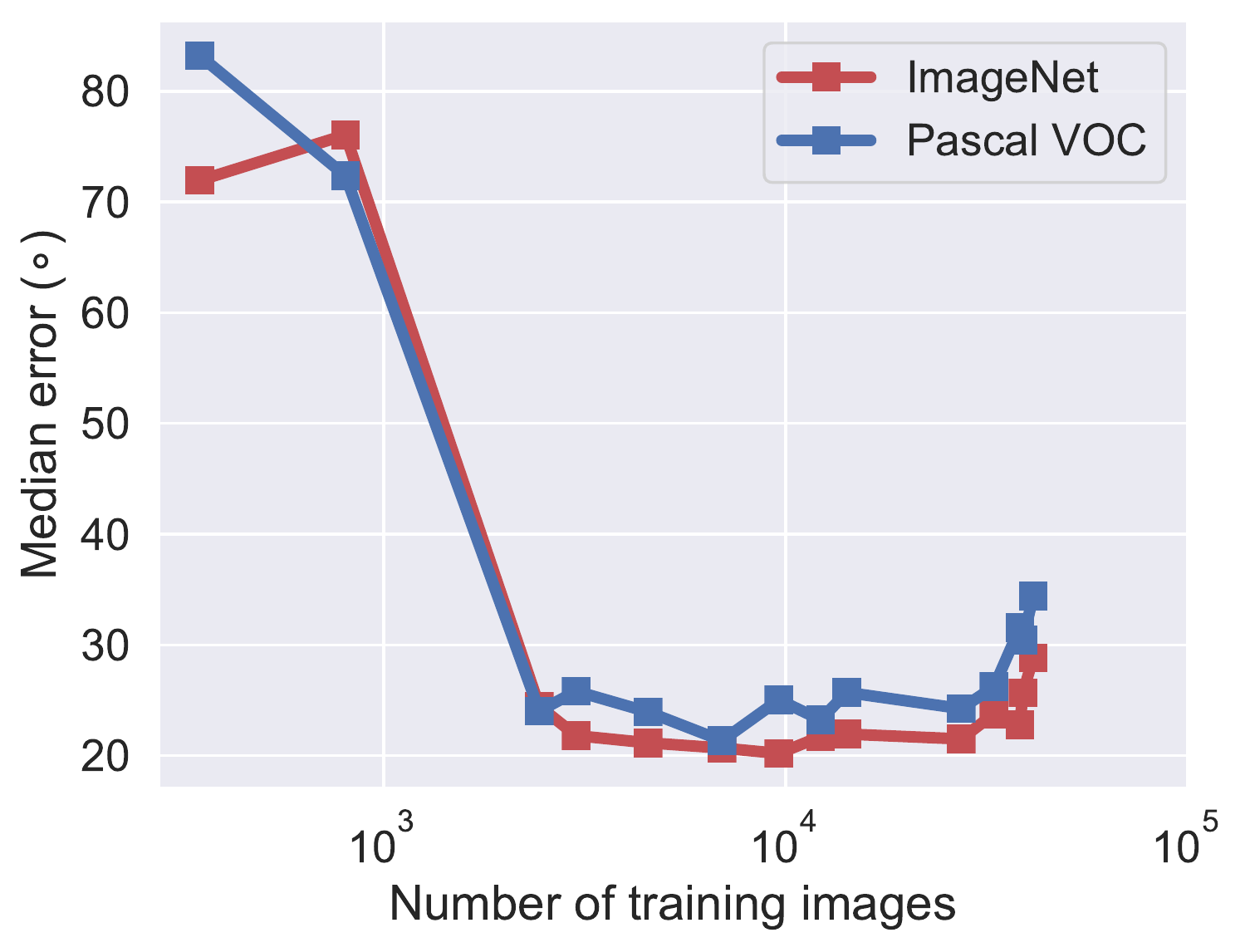} \\
    \multicolumn{2}{c}{\textbf{Random initialization}} \\
    \end{tabular}
    \caption{\textbf{The effect of annotation noise level on 3D pose prediction is consistent across different network initialization.} 
    For each initialization method, we report the performance of the pose predictor under different noise level of pose annotations.
    Higher level of annotation noise corresponds to larger number of training images. 
    We report both prediction accuracy (top row) and median error (bottom row) on two test splits included in PASCAL3D+ (\ie PASCAL VOC and ImageNet validation set.).
    }
    \label{fig:noise-level-init}
    \vspace{5mm}
\end{figure}

\paragraph{The effect of annotation noise level on pose estimation}\label{sec:init}
In the main text, we use ImageNet-pretrained ResNet50 to initialize our model and analyze the effect of annotation noise level to the performance of pose estimation (Fig.~6 in the main paper). 
Here we provide additional experimental results under different network initialization including contrastively pretraining and randomly initialization. 
Fig.~\ref{fig:noise-level-init} demonstrates that the effect of annotation noise level on the pose estimation performance is consistent across different network initialization, \ie neither clean-yet-small data nor large-yet-noisy data leads to higher performance than mid-size data with mid-level noise.

\section{Implementation}\label{sec:more-implementation}

We use the Structure-from-Motion (SfM) and Multiview Stereo (MVS) pipelines implemented in COLMAP~\cite{schoenberger2016sfm,schoenberger2016mvs}~\footnote{\url{https://colmap.github.io/}} and HLOC library~\cite{sarlin2019coarse}~\footnote{\url{https://github.com/cvg/Hierarchical-Localization}}.
We use the MaskRCNN~\cite{he2017mask} implemented in Detectron2~\cite{wu2019detectron2} to get the object masks.
We implemente our pose estimation models based on PoseContrast~\cite{xiao2021posecontrast}~\footnote{\url{https://github.com/YoungXIAO13/PoseContrast}}.

%%%%%%%%% REFERENCES
{\small
\bibliographystyle{ieee_fullname}
\bibliography{egbib}
}
\end{document}